%% file: main.tex
\pdfoutput=1
\documentclass[11pt]{article}
\usepackage[final]{acl}
\usepackage{subfiles}

\usepackage{times}
\usepackage{latexsym}
\usepackage{amsmath} % Hỗ trợ một số biểu thức toán học
\usepackage{amssymb} % Bổ sung thêm kí hiệu về toán học
\usepackage{amsbsy} % Hỗ trợ các kí hiệu in đậm
\usepackage{booktabs}
\usepackage[T1]{fontenc}
\usepackage[utf8]{inputenc}

\usepackage{microtype}
\usepackage{todonotes}
\usepackage{inconsolata}

\usepackage{graphicx}
\usepackage{subcaption}
\usepackage{multirow}
\usepackage{array}
\usepackage{longtable}
\usepackage{supertabular}

\title{Who's Who: Large Language Models Meet Knowledge Conflicts in Practice}

\author{Quang Hieu Pham$^{1*}$, Hoang Ngo$^{1*}$, Anh Tuan Luu$^2$, Dat Quoc Nguyen$^1$ \\
         $^1$VinAI Research, Vietnam; $^2$Nanyang Technological University, Singapore\\
         $^1$\texttt{\{v.hieupq1, v.hoangnv49, v.datnq9\}@vinai.io}; $^2$\texttt{anhtuan.luu@ntu.edu.sg}}

\begin{document}
 \maketitle
 \def\thefootnote{*}\footnotetext{The first two authors contributed equally to this work.}\def\thefootnote{\arabic{footnote}}

\begin{abstract}
\subfile{Sections/Abstract.tex}
\end{abstract}

\section{Introduction}
\subfile{Sections/1_Introduction.tex}

\section{Our WhoQA Dataset}
\subfile{Sections/2_OurWhoQA.tex}

\section{Experiments}
\label{sec:5 experiment results}
\subfile{Sections/5_Experiment.tex}

\section{Conclusions}
\subfile{Sections/Conclusion.tex}

\section*{Limitations}
\subfile{Sections/Limitations.tex}
% \section*{Acknowledgments}

\bibliography{custom}

\newpage 

\appendix

\label{sec:appendix}
\subfile{Sections/Appendix}

\end{document}

%% file: Sections/Abstract.tex
Retrieval-augmented generation (RAG) methods are viable solutions for addressing the static memory limits of pre-trained language models. Nevertheless, encountering conflicting sources of information within the retrieval context is an inevitable practical challenge. In such situations, the language models are recommended to transparently inform users about the conflicts rather than autonomously deciding what to present based on their inherent biases. To analyze how current large language models (LLMs) align with our recommendation, we introduce WhoQA, a public benchmark dataset to examine model's behavior in knowledge conflict situations. We induce conflicts by asking about a common property among entities having the same name, resulting in questions with up to 8 distinctive answers. WhoQA evaluation set includes 5K questions across 13 Wikidata property types and 150K Wikipedia entities. Our experiments show that despite the simplicity of WhoQA questions, knowledge conflicts significantly degrades LLMs' performance in RAG settings.

%% file: Sections/1_Introduction.tex
Recent large-scale pretrained language models (LLMs) excel in tasks requiring natural language understanding \citep{radford2019language, brown2020language, chowdhery2023palm, Achiam2023GPT4TR}. However, they often "hallucinate" plausible but incorrect content due to outdated or incorrect pretraining information \citep{parikh-etal-2020-totto, wang-2019-revisiting, dhingra-etal-2022-time, luu-etal-2022-time}. Retrieval augmented generation (RAG) methods provide contextual knowledge to address this, but conflicts within retrieval contexts are still inevitable \citep{chen2024benchmarking, ge2024time}.

\begin{figure}[t]
  \includegraphics[width=\columnwidth]{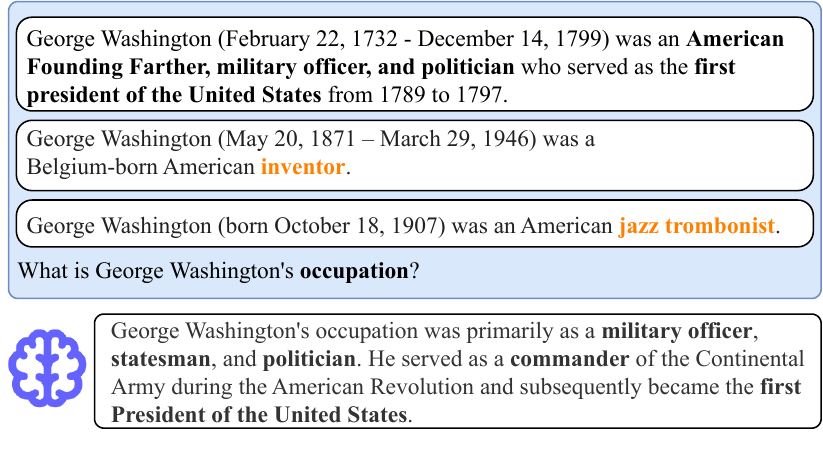} % fine
  \caption{An example of when an LLM inherently biases an entity over another entity.}
  \label{fig:conflict-demo}
\end{figure}

According to \citet{xie2024adaptive}, under certain knowledge conflict circumstances, large language models (LLMs) may either ignore contextual knowledge and rely solely on their parametric memory or, more concerning, prefer certain contextual knowledge based on their popularity or specific input ordering. Figure \ref{fig:conflict-demo} illustrates a conflict where an LLM prioritizes the popular facts about George Washington, the U.S. president, over contexts about an inventor and a jazz trombonist with the same name. This action can lead to information loss and bias, particularly when users have limited knowledge about the subject and are conducting an information-seeking query. Since eliciting LLMs parametric knowledge to handle specific cases is costly \citep{xie2024adaptive}, we recommend that LLMs should inform their users about conflicting information, provide proper citations, and allow users to make informed decisions based on the presented evidence.

To examine if current LLMs handle knowledge conflicts transparently and accurately, we construct WhoQA, a dataset derived from Wikipedia articles and their linked Wikidata entities \cite{denny2014wikidata}. In WhoQA, we intentionally induce conflicts by making questions asking about shared properties of Wikipedia entities with the same name. Previous studies have explored LLMs' behavior in knowledge conflicts, either by substituting named entities \cite{longpre-etal-2021-entity,neeman-etal-2023-disentqa} or using another LLM for counterfactuals \cite{xie2024adaptive}. These methods can generate non-existent facts, which LLMs, trained for factual consistency, may avoid. Our work mends the gap by sourcing all supporting documents from Wikipedia, a widely used and trusted source for many NLP tasks. With 2-8 different answers per question, compared to a maximum of two in previous works \cite{longpre-etal-2021-entity, neeman-etal-2023-disentqa, xie2024adaptive}, WhoQA demonstrates the complication of knowledge conflicts, which are not only natural but also prevalent in practice.

Similar to solving knowledge conflicts, AmbigQA \cite{min2020ambigqa} rewrites ambiguous user queries so that each version has only one answer. However, sometimes there can be no information to resolve the ambiguity. For example, the statements "George Washington is a farmer" and "George Washington is the first president of the United States" both refer to the same person, but can seem contradictory, if found in two separate documents without context. SituatedQA \cite{zhang2021situatedqa} creates knowledge conflicts by considering similar questions in different temporal or geographical contexts. In WhoQA, where many entities share the same name, a wider range of properties can induce knowledge conflicts.

In summary, our contributions include:

\begin{itemize}
    \item We highlight the prevalence of knowledge conflicts in practical settings and manually construct an evaluation set of 5,152 questions with supporting evidence, serving as a gold-standard benchmark. 
    \item We perform extensive experiments using strong LLMs and show that knowledge conflicts pose a significant challenge, potentially misleading or biasing LLM users.
    \item  We publicly release WhoQA to foster future research of knowledge conflicts in LLMs. Our WhoQA is available at: \url{https://github.com/VinAIResearch/WhoQA}.
\end{itemize}

%% file: Sections/2_OurWhoQA.tex
From the set of named entities in the English Wikipedia and Wikidata dump, we group all entities that have the same name together, collect question-answer pairs as well as supporting contexts, and manually revise the evaluation set. 

Take an example from Figure \ref{fig:conflict-demo} with a set $S$ containing three people whose names are George Washington: one is the first President of the US, who was also a military officer; another one is a Belgium-born American inventor; and the last one is an American jazz trombonist. Given a question asking about a property $p = \text{``occupation''}$  of George Washington---$q = ``\textit{What is George Washington's occupation?}"$ and a set $C$ of supporting contexts from all the three George Washington's Wikipedia documents, a model is expected to give all answers within a set $A =$ \{\{President of the US, military officer\}, \{inventor\}, \{jazz trombonist\}\}.

Formally, WhoQA is a multi-answer question answering dataset in which each question can be described as a quadruplet $(q, A, S, C)$, where $q$ is a question asking about a property $p$ of a named-entity $s$;\ \ $A = \{A_i\}^m_{i=1}$ is a set of $m$ possible distinctive answer sets to $q$;\ \ $S = \{s_i\}^n_{i=1}$ is a set of $n$ named entities that share the same name mentioned in $q$; and $C = \{c_i\}^n_{i=1}$ where $c_i$ is the corresponding supporting context for entity $s_i$. 
In a real-world RAG setup, $C$ is obtained from a retrieval system. In this work, we focus on studying knowledge conflicts, hence each $c_i$ is a textual context extracted from the corresponding Wikipedia document of $s_i$. 
\paragraph{Question-Answer pairs collection:} Starting with a set $S$ containing named entities $s_i$ that share the same name, we induce knowledge conflicts by creating questions about all shared properties (e.g., occupation, date of birth, and the like) among the entities. We only consider the top 60 most popular properties reported by Wikipedia.
For each property, we manually construct 5 question templates and substitute the shared name among entities in $S$ to create specific questions. An example of a template for the property $p = \text{``occupation''}$ is \textit{``What is <shared\_entity\_name>'s occupation?''}. See details of these question templates in Appendix \ref{appendix:qtemplates}. 

For each entity $s_i$ in the set $S$, the answer to a question $q$, created for the property $p$, is the corresponding set $A_i$ in the Wikidata property triplet $(s_i, p, A_i)$ of $s_i$. 
We only consider properties $p$ in which the corresponding answer set $A_i$ contains Wikidata entities, plain texts or a date-time string. If $A_i$ is a Wikipedia entity, we take all the alternative names annotated by Wikidata. For $A_i$ that are date-time strings, we follow the Wikipedia Manual of Style instructions to cover all of their recommended written forms. We left $A_i$ which are plain texts in their original form. 
This step generates about 293K questions, not counting different template variations, highlighting the prevalence of knowledge conflicts in practice. 
\paragraph{Supporting context collection:} For the corresponding Wikipedia document of each entity $s_i$  in $S$, we search for all occurrences of elements within the corresponding set $A_i$ and take contexts of 256 tokens around the most frequently occurring element as candidate supporting contexts. Then, we concatenate the question $q$ with the most frequently occurring element, and use Contriever \cite{izacard2021contriever} to calculate candidate contexts' relevant score. The final score for each context is averaged over the 5 questions generated from our 5 question templates. We keep only the most relevant context, if its score is above 0.45. Otherwise, the Wikipedia document is removed from our dataset. Finally, we filter the set $A_i$ to include only elements that appear together with the most frequent element in the top-1 context. Our simple matching process typically results in supporting contexts where the answers are directly stated.

\paragraph{Final Benchmark:}
After the filtering steps above, 76,487 questions remain, spanning 13 property types and involving 145,710 different Wikidata entities. To study the effect of property types and the number of conflicting answers on LLM performance, we undersample properties with many questions and prioritize those with many conflicting answers. For each property, we randomly select up to 500 questions, using weighted sampling where questions with more conflicting answers have a higher probability of being selected.
To ensure the reliability of our evaluation set, we hire two annotators to independently verify whether the answer elements in $A_i$ can be inferred from the automatically collected supporting contexts. Initially, the annotators have to check 18,967 contexts for 5,592 questions. However, we allow them to exclude a question, if there is at least one incorrectly labeled context. In total, the annotators exclude 1,602 contexts together with their corresponding questions and answers. Our post checking process finally returns an evaluation set containing 5,152 questions, with 3.25 conflicting answers per question on average and $18,967-1,602 = 17,365$ contexts.

\begin{figure}[t!]
  \includegraphics[width=\columnwidth]{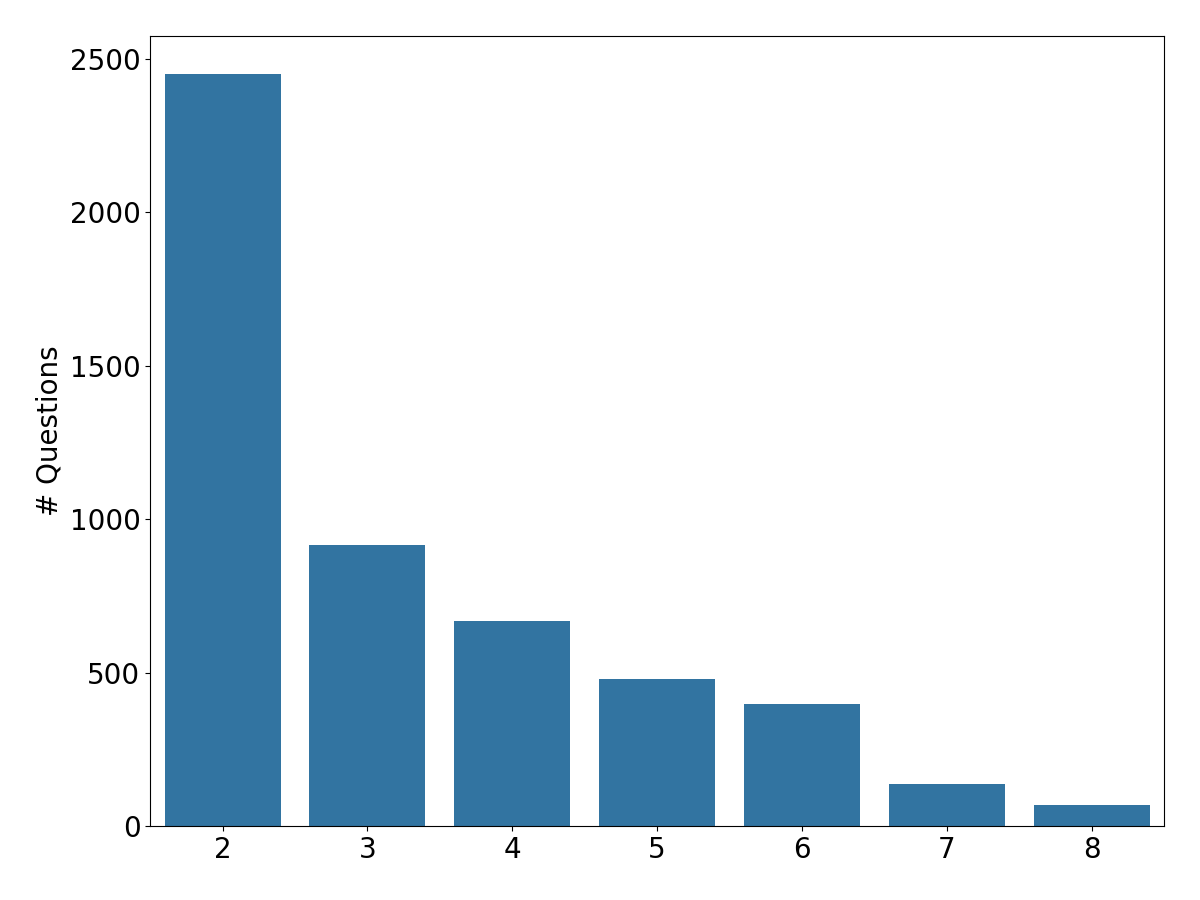}
  \caption{Number of questions distribution with respect to number of conflicting answers}
  \label{fig:eval_stats_ans}
\end{figure}

\begin{figure}[t!]
  \includegraphics[width=\columnwidth]{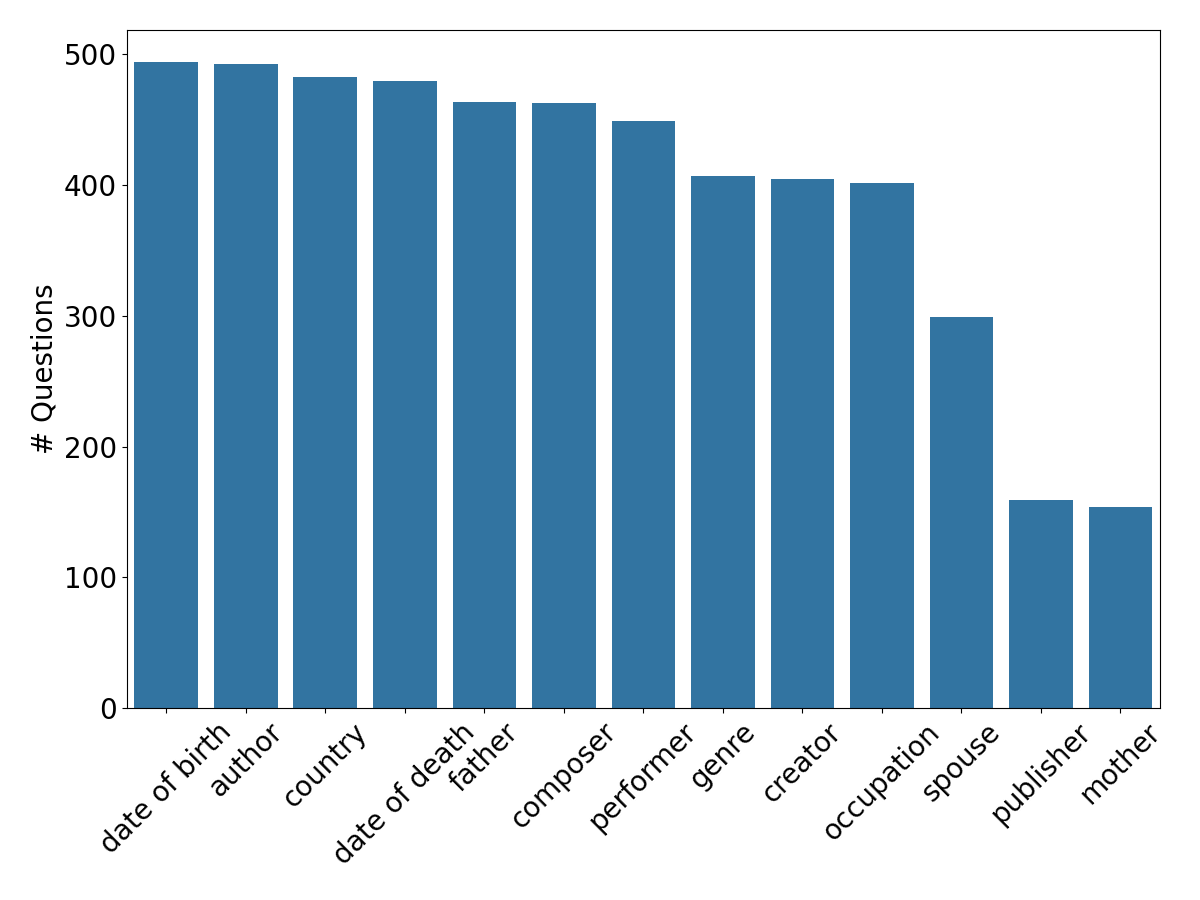}
  \caption{Number of question with respect to questions' property types}
  \label{fig:eval_stats_rel}
\end{figure}

Figure \ref{fig:eval_stats_ans} and Figure \ref{fig:eval_stats_rel} demonstrates the overall statistics of our dataset with respect to the number of conflicting answers and types of Wikidata property respectively.

%% file: Sections/5_Experiment.tex
\subsection{Experimental setup}
We study LLMs' behavior when dealing with knowledge conflicts in three different scenarios:

\begin{enumerate}
    \item The first setting is a simple QA scenario where only a single context is presented to a model, hence no conflict.
    \item In the second setting, we provide the model with conflicting contexts without informing it about the presence of conflicts.
    \item Finally, in the third setting, we explicitly specify the presence of conflicts in the model's input.
\end{enumerate}

We evaluate 10 open source models: Gemma 7B \cite{team2024gemma}, Mistral 7B \cite{jiang2023mistral} , Mixtral 8x7B \cite{jiang2024mixtral}, Qwen1.5 Chat (7B, 14B, 32B, 72B) \cite{qwen}, Command R 35B \cite{commandr}, and  Llama 3 (8B, 70B) \cite{llama3modelcard}. We also evaluate GPT-3.5 \cite{ouyang2022training} behavior via OpenAI API. See our prompt templates for all these experimental models in Appendix \ref{appendix:prompt_templates}.

We serve all the open source LLMs in our experiments using vLLM \cite{kwon2023efficient}, with bfloat16 floating-point format on a single machine with 8 Nvidia A100 40GB GPUs. The inference process uses sampling with $\text{top\_k}=50$ tokens, $\text{top\_p}=0.9$, $\text{max\_tokens} = 512$, and $\text{temperature}=0.9$.

\paragraph{Accuracy evaluation metric:} 
An answer is correct if it includes all references to the elements in the answer set supported by our collected contexts. As the Wikidata database already contains various name variations for its entities, this significantly reduces the chance of missing correct answers from the models' responses.

\subsection{Simple QA} \label{sec:simple-qa}

\begin{table}[t!]
\centering
\resizebox{\columnwidth}{!}{
\begin{tabular}{@{}lcccc@{}}
\toprule
Model & \#Param & \multicolumn{1}{l}{SimQA} & \multicolumn{1}{l}{W/oS} & \multicolumn{1}{l}{W/S} \\ \midrule
Gemma 1.1    & 7B                   & 91.8                 & 12.7               & 34.4                 \\
Mistral      & 7B                   & 96.5                 & 44.5                & 54.7                 \\
Qwen1.5 Chat & 7B                   & 91.9                 & 22.0                & 50.2                 \\
Llama 3      & 8B                   & 98.2                 & 72.4                 & 71.5                 \\
Qwen1.5 Chat & 14B                  & 95.5                 & 9.90                & 51.9                 \\
Qwen1.5 Chat & 32B                  & 95.5                 & 21.0                 & 60.5                 \\
Command R    & 35B                  & 92.5                 & 69.9                 & -                  \\
Mixtral 8x7B & 40B                  & 97.1                 & 53.7                & 64.2                 \\
Llama 3      & 70B                  & 97.8                 & 83.8                & 86.1                 \\
Qwen1.5 Chat & 72B                  & 95.9                 & 28.2               & 66.8                 \\ \midrule
gpt-3.5-turbo      & - & 97.9 & 36.0 & 58.9 \\ \bottomrule
\end{tabular}
}
\caption{Macro average accuracy of all models over the number of conflicting answers. ``SimQA'', ``W/oS'' and ``W/S'' denote the first setting of simple QA, the second setting of knowledge conflict without specification and the third setting of knowledge conflict with specification, respectively. Due to the chat template of Command R, it cannot be used for the third setting ``W/S''. } 
\label{tab:general results}
\end{table}

We divide each question into $n$ turns, each corresponding to one of the $n$ conflicting contexts. As shown in Table \ref{tab:general results}, when conflicts are removed, most LLMs can accurately answer single-hop questions. This result highlights the ease of finding answers within the individual contexts of WhoQA, allowing our subsequent experiments to focus more on the issue of knowledge conflicts.

\subsection{Knowledge conflict: without specification}
We hypothesize that an ideal LLM should inherently recognize knowledge conflicts. To test this, we provide all conflicting contexts for each question and ask the LLM to answer without indicating the presence of conflicts. 

\paragraph{Simple knowledge conflicts substantially impairs LLMs.} As shown in Table \ref{tab:general results}, LLMs in our experiments exhibit significant performance drops when knowledge conflicts occur. Since finding answers from each single context in WhoQA is straightforward, this decline is attributable to knowledge conflicts. Therefore, the varying levels of performance drop indicate LLMs' robustness against knowledge conflicts.

\paragraph{Without being noticed, LLMs are not sensitive to subtle knowledge conflicts.} We examine how LLM performance changes as the number of conflicting answers increases. It is logical to assume that more conflicting information in the input context leads to higher information entropy. Table \ref{tab:acc-by-num-ans} shows that %although there is a significant performance gap compared to settings with no knowledge conflict, 
LLMs generally perform better on questions with more conflicting answers. This suggests that LLMs are less sensitive to subtle conflicts within their input contexts. Therefore, LLM practitioners should address knowledge conflicts in tasks where fine-grained answers are expected.

\begin{table}[t!]
\resizebox{\columnwidth}{!}{%
\begin{tabular}{@{}l|c|cc|cc@{}}
\toprule
Model & \#Param & \multicolumn{1}{l}{$\leq$ 4 Wo/S} & {$>$ 4 Wo/S} & $\leq$ 4 W/S & $>$ 4 W/S \\ \midrule
Gemma 1.1                         & 7B  & \textbf{16.5} & 9.80          & \textbf{41.7} & 29.0 \\
Mistral                           & 7B  & \textbf{53.0} & 38.1          & \textbf{59.9} & 50.7 \\
Qwen1.5 Chat                      & 7B  & 20.9 & \textbf{22.9} & \textbf{56.8} & 45.3 \\
Llama 3                           & 8B  & 70.6 & \textbf{73.7} & \textbf{76.2} & 67.9 \\
Qwen1.5 Chat                      & 14B & 5.50 & \textbf{13.2} & 49.4          & \textbf{53.8} \\
Qwen1.5 Chat                      & 32B & 11.7 & \textbf{28.0} & \textbf{61.8} & 59.6 \\
Command R                         & 35B & \textbf{77.9} & 63.9          & -              & -     \\
Mixtral 8x7B                      & 40B & 51.5 & \textbf{55.3} & \textbf{64.8} & 63.7 \\
Llama 3                           & 70B & 83.8 & \textbf{83.8} & \textbf{87.9} & 84.7 \\
Qwen1.5 Chat                      & 72B & 25.0 & \textbf{30.6} & 60.9          & \textbf{71.3} \\
gpt-3.5-turbo                     & -   & \textbf{45.8} & 28.6          & 51.8          & \textbf{64.2} \\ \bottomrule
\end{tabular}
}
\caption{ Average model performance variation across different number of conflicting answers ($\leq$ 4 or $>$ 4) in the input context.}
\label{tab:acc-by-num-ans}
\end{table}

\paragraph{There is a tradeoff between helpfulness and accuracy when LLMs meet knowledge conflicts.} 
We consider questions where LLMs miss all answers from all contexts as hard questions and manually review at most 100 hard question responses for each LLM. Table \ref{tab:rejection-rate} shows the rate at which each model gives no answer to questions. Specifically, gpt-3.5-turbo often states that there are multiple individuals with the same name in the input contexts and asks for clarification. While safe, this kind of response requires the user to read all contexts to clarify, providing no direct benefit. Gemma 7B, instead of acknowledging conflicts, simply states there is not enough information to answer, which can confuse the user. For models with low rejection rates like LLama3 70B, Qwen1.5 72B, and Mixtral 8x7B, handling conflicts is crucial. Reviewing Qwen1.5 72B responses, we find that Qwen often selects only one answer among many. This behavior is more concerning than refraining from answering, as it can cause misinformation and bias the users.

\begin{table}[t!]
\begin{tabular}{@{}lcc@{}}
\toprule
Model            & \multicolumn{1}{l}{\#Param} & \multicolumn{1}{l}{Rejection rate (\%)} \\ \midrule
gpt-3.5-turbo & -                             & 99                                      \\
Qwen1.5-Chat     & 72B                           & 8                                       \\
LLama 3          & 70B                           & 33                                      \\
Mixtral 8x7B     & 40B                           & 18                                      \\
Gemma 1.1        & 7B                            & 92                                      \\ \bottomrule
\end{tabular}
\caption{Rejection to answer rate of models on their own hard evaluation set. See the full result table with all models in Appendix \ref{appendix:rejectionrate}.}
\label{tab:rejection-rate}
\end{table}

\subsection{Knowledge conflict: with specification}
We modify the prompt template to include few-shot examples, notifying LLMs of knowledge conflicts within their input context.
Results in Table \ref{tab:general results} show significant performance improvement for most models. The exceptions are the two LLama3 models, which show minimal benefit.  A review of their responses reveals that these models are aware of conflicts and attempt to provide all available answers, regardless of whether conflicts are stated. This behavior supports our view that LLMs should be transparent and serve as tools to aid users' decisions. Results from Table \ref{tab:acc-by-num-ans} suggest that questions with more conflicting answers are indeed more challenging for LLMs.

%% file: Sections/Conclusion.tex
We introduce WhoQA, a high-quality benchmark dataset designed to examine language model behaviors in knowledge conflict situations. WhoQA bridges the gap with previous work by inducing knowledge conflicts without generating counterfactuals, demonstrating that conflicts can arise not only from misinformation but also naturally due to ambiguity in retrieved contexts, user questions, or similarities among entities. Through our experiments, we also show that many LLMs are not sensitive to subtle conflicts within their input, and thus only simple conflicts can significantly impair LLM performance in RAG settings.

%% file: Sections/Limitations.tex
The simplicity of our matching process to collect supporting contexts results in the fact that questions in WhoQA are mostly single-hop. We think that it could be noteworthy to propose a way to induce knowledge conflicts for multi-hop questions so as to challenge future LLMs. We still leave the question on how to make LLMs' responses more informative in case on dealing with knowledge conflict. Possibly, looking at fine-tuning methods which control how LLMs deal with unfamiliar examples \cite{kang2024unfamiliar,ge2024time} is a promising direction.

%% file: Sections/Appendix.tex
\section{Question templates}\label{appendix:qtemplates}
We include in Table \ref{tab:qtemplates} all the 65 question templates for the final 13 Wikidata properties in our final dataset.

% \usepackage{booktabs}
% \usepackage{graphicx}
\begin{table*}[t!]
\resizebox{0.4\paperheight}{!}{%
\begin{tabular}{@{}l|l@{}}
\toprule
Property &
  Question template \\ \midrule
date of birth &
  \begin{tabular}[c]{@{}l@{}}1. What is the date of birth of {[}subject{]}?\\ 2. When was {[}subject{]} born?\\ 3. What is the birthdate of {[}subject{]}?\\ 4. What is the date of {[}subject{]}'s birth?\\ 5. When did {[}subject{]} come into this world?\end{tabular} \\ \midrule
author &
  \begin{tabular}[c]{@{}l@{}}1. What is the author of {[}subject{]}?\\ 2. Who created {[}subject{]}?\\ 3. What is the name of the main creator of {[}subject{]}?\\ 4. Who wrote {[}subject{]}?\\ 5. Can you provide the authorship of {[}subject{]}?\end{tabular} \\ \midrule
country &
  \begin{tabular}[c]{@{}l@{}}1. What is the country of origin for {[}subject{]}?\\ 2. Which country does {[}subject{]} reside in?\\ 3. Can you tell me the country where {[}subject{]} is located?\\ 4. What is the sovereign state where {[}subject{]} is situated?\\ 5. What country does {[}subject{]} belong to?\end{tabular} \\ \midrule
date of death &
  \begin{tabular}[c]{@{}l@{}}1. What is the date of death of {[}subject{]}?\\ 2. When did {[}subject{]} pass away?\\ 3. What was the date of {[}subject{]}'s demise?\\ 4. When did {[}subject{]} die?\\ 5. Can you tell me the date of {[}subject{]}'s passing?\end{tabular} \\ \midrule
farther &
  \begin{tabular}[c]{@{}l@{}}1. Can you tell me who is the farther of {[}subject{]}?\\ 2. Who is the father of {[}subject{]}?\\ 3. What is {[}subject{]}'s farther name?\\ 4. Could you give me the name of {[}subject{]}'s farther?\\ 5. Tell me who is {[}subject{]}'s farther?\end{tabular} \\ \midrule
composer &
  \begin{tabular}[c]{@{}l@{}}1. Who is the composer of {[}subject{]}'s music?\\ 2. Can you tell me who wrote the music for {[}subject{]}?\\ 3. Who are the individuals responsible for composing {[}subject{]}'s music?\\ 4. Could you specify the composer of {[}subject{]}?\\ 5. Who compose the music for {[}subject{]}?\end{tabular} \\ \midrule
performer &
  \begin{tabular}[c]{@{}l@{}}1. Who is the performer associated with {[}subject{]}?\\ 2. Could you please identify the individual or group acting as the performer for {[}subject{]}?\\ 3. Who takes on the role of performer of {[}subject{]}?\\ 4. Who is the performer who perform {[}subject{]}?\\ 5. Who performs {[}subject{]}?\end{tabular} \\ \midrule
genre &
  \begin{tabular}[c]{@{}l@{}}1. What genre does {[}subject{]} belong to?\\ 2. Could you specify the genre of {[}subject{]}?\\ 3. What genre would {[}subject{]} be classified as?\\ 4. Which genre would you classify {[}subject{]} into?\\ 5. Which genre would {[}subject{]} be classified into?\end{tabular} \\ \midrule
creator &
  \begin{tabular}[c]{@{}l@{}}1. Who is attributed as the creator of the {[}subject{]}?\\ 2. By whom was the {[}subject{]} created?\\ 3. Who is responsible for making the {[}subject{]}?\\ 4. Who can be identified as the creator of the {[}subject{]}?\\ 5. What individual or entity is credited with creating the {[}subject{]}?\end{tabular} \\ \midrule
occupation &
  \begin{tabular}[c]{@{}l@{}}1. What is the {[}subject{]}'s occupation?\\ 2. Could you tell me what {[}subject{]} does for a living?\\ 3. What profession or job is associated with {[}subject{]}?\\ 4. What profession does {[}subject{]} work in?\\ 5. What role or position does {[}subject{]} hold professionally?\end{tabular} \\ \midrule
spouse &
  \begin{tabular}[c]{@{}l@{}}1. Can you tell me who {[}subject{]}'s spouse is?\\ 2. Who is married to {[}subject{]}?\\ 3. What is the name of {[}subject{]}'s spouse?\\ 4. Who is {[}subject{]} spouse?\\ 5. Who is the spouse of {[}subject{]}?\end{tabular} \\ \midrule
publisher &
  \begin{tabular}[c]{@{}l@{}}1. Who is the publisher of the {[}subject{]}?\\ 2. Which organization or person is responsible for publishing the {[}subject{]}?\\ 3. Who is responsible for bringing the {[}subject{]} to the public?\\ 4. Can you name the publisher of the {[}subject{]}?\\ 5. Is there any information on who published the {[}subject{]}?\end{tabular} \\ \midrule
mother &
  \begin{tabular}[c]{@{}l@{}}1. Can you tell me who is the mother of {[}subject{]}?\\ 2. Who is the mother of {[}subject{]}?\\ 3. What is {[}subject{]}'s mother name?\\ 4. Could you give me the name of {[}subject{]}'s mother?\\ 5. Tell me who is {[}subject{]}'s mother?\end{tabular} \\ \bottomrule
\end{tabular}%
}
\caption{Question templates for properties in WhoQA, sorted descending based on the number of question per each property in the dataset. Here [subject] denotes the <shared\_entity\_name> within a set of entities $S$.}
\label{tab:qtemplates}
\end{table*}

\section{Prompt templates}\label{appendix:prompt_templates}
We report the prompt templates used in our experiments in Table \ref{tab:prompt-templates}.

\begin{table*}[t!]
\resizebox{\textwidth}{!}{%
\begin{tabular}{p{0.25\textwidth}|p{0.75\textwidth}}
\hline
\textbf{Setting} &
  \textbf{Prompt template} \\ \hline
\textbf{Conflict without specification} &
  \begin{tabular}[c]{@{}p{0.75\textwidth}@{}}Given the following documents, answer the question under the \#\#\# QUESTION section. Give short answer only.\\ \\ \#\#\# DOCUMENTS:\\ \{docs\}\\ \\ \#\#\# QUESTION:\\ \{question\}\\ \\ \#\#\# RESPONSE:\end{tabular} \\ \hline
\textbf{Conflict with specification} &
  \begin{tabular}[c]{@{}p{0.75\textwidth}@{}}\#\#\# INSTRUCTIONS\\ 1. You are given some DOCUMENTS that are relevant to answer a QUESTION \\ 2. Read the DOCUMENTS, think step-by-step and give your final answer(s) to the question. \\ 3. Give your answer(s) as a list with each item starting with "-". Do not include any other formatting. In your answer(s), give only the necessary information in a concise format. Here are some examples for you to learn from. \\ \\ Example 1:\\ \#\#\# DOCUMENTS:\\ Doc \#0:\\ Foo was written by Bar\\ \\ Doc \#1:\\ Foo was written by Boo\\ \\ \#\#\# QUESTION:\\ Who is the author of Foo?\\ \\ \#\#\# RESPONSE:\\ - Bar (Doc \#0)\\ - Boo (Doc \#1)\\ \\ Example 2:\\ \#\#\# DOCUMENTS:\\ Doc \#0:\\ Bar met Foo in {[}year1{]}\\ \\ Doc \#1:\\ Foo met Bar in {[}year2{]}\\ \\ \#\#\# QUESTION:\\ When did Foo meet Bar?\\ \\ \#\#\# RESPONSE:\\ - {[}year1{]} (Doc \#0)\\ - {[}year2{]} (Doc \#1)\\ \\ Your task:\\ \#\#\# DOCUMENTS:\\ \{docs\}\\ \\ \#\#\# QUESTION:\\ \{question\}\\ \\ \#\#\# RESPONSE:\end{tabular} \\ \hline
\end{tabular}%
}
\caption{Prompt templates.}
\label{tab:prompt-templates}
\end{table*}

\section{Models rejection rate}\label{appendix:rejectionrate}
We consider questions where LLMs miss all answers from all contexts as hard questions and manually review 100 hard question responses for each LLM. Table \ref{tab:full-rejection-rate} shows the rate at which each model gives no answer to questions.

\begin{table}[t!]
\resizebox{\columnwidth}{!}{%
\begin{tabular}{lcc}
\hline
Model         & \multicolumn{1}{l}{\#Param} & \multicolumn{1}{l}{Rejection rate (\%)} \\ \hline
gpt-3.5-turbo & -                           & 99                                      \\
Qwen1.5-Chat  & 72B                         & 8                                       \\
LLama 3       & 70B                         & 33                                      \\
Mixtral 8x7B  & 40B                         & 18                                      \\
Qwen1.5-Chat  & 32B                         & 2                                       \\
Qwen1.5-Chat  & 14B                         & 0                                       \\
Qwen1.5-Chat  & 7B                          & 0                                       \\
Mistral       & 7B                          & 47                                      \\
Llama 3       & 8B                          & 23                                      \\
Gemma 1.1     & 7B                          & 92                                      \\ \hline
\end{tabular}%
}
\caption{Rejection to answer rate of models on their own hard evaluation set.}
\label{tab:full-rejection-rate}
\end{table}